\documentclass[sigconf]{acmart}

\AtBeginDocument{%
  \providecommand\BibTeX{{%
    \normalfont B\kern-0.5em{\scshape i\kern-0.25em b}\kern-0.8em\TeX}}}

\copyrightyear{2023}
\acmYear{2023}
\setcopyright{acmlicensed}\acmConference[DocIU Workshop, CIKM '23]{Proceedings of the 32nd ACM International Conference on Information and Knowledge Management}{October 21--25, 2023}{Birmingham, United Kingdom} \acmBooktitle{Proceedings of the 32nd ACM International Conference on Information and Knowledge Management (DocIU Workshop, CIKM '23), October 21--25, 2023, Birmingham, United Kingdom}
\acmDOI{https://doi.org/XXXXXX.XXXX}

\begin{document}

\title{Enhancing BERT-Based Visual Question Answering through Keyword-Driven Sentence Selection}

\author{Davide Napolitano}
\authornote{Both authors contributed equally to this research.}
\email{davide.napolitano@polito.it}
\orcid{0000-0001-9077-4103}
\affiliation{%
  \institution{Politecnico di Torino}
  \city{Turin}
  \country{Italy}
}

\author{Lorenzo Vaiani}
\email{lorenzo.vaiani@polito.it}
\authornotemark[1]
\orcid{0000-0002-3605-1577}
\affiliation{%
  \institution{Politecnico di Torino}
  \city{Turin}
  \country{Italy}
}

\author{Luca Cagliero}
\email{luca.cagliero@polito.it}
\orcid{0000-0002-7185-5247}
\affiliation{%
  \institution{Politecnico di Torino}
  \city{Turin}
  \country{Italy}
}

\begin{abstract}
The Document-based Visual Question Answering competition addresses the automatic detection of parent-child relationships between elements in multi-page documents.
The goal is to identify the document elements that answer a specific question posed in natural language. 
This paper describes the PoliTo's approach to addressing this task, in particular, our best solution explores a text-only approach, leveraging an ad hoc sampling strategy.
Specifically, our approach leverages the Masked Language Modeling technique to fine-tune a BERT model, focusing on sentences containing sensitive keywords that also occur in the questions, such as references to tables or images.
Thanks to the effectiveness of this approach, we are able to achieve high performance compared to baselines, demonstrating how our solution contributes positively to this task.

\end{abstract}

\ccsdesc[500]{Computing methodologies~Natural language processing}
\ccsdesc[500]{Computing methodologies~Computer vision}
\ccsdesc[500]{Computing methodologies~Machine learning}

\keywords{Visual Question Answering, Document Understanding}



\maketitle

\section{Introduction}

In the realm of natural language understanding and document analysis, the intersection of these fields has given rise to a captivating challenge: the development of a sophisticated model capable of processing both linguistic queries and collections of Regions of Interest (RoIs) within documents. This challenge hinges on the pivotal task of accurately predicting the index or indices of the RoIs that hold the key to answering the posed questions.

The amalgamation of these components demands a multidisciplinary approach, combining advanced natural language processing with document layout comprehension. In this report, we delve into our solution, which centers on harnessing the power of a BERT model~\cite{bert} fine-tuned on a subset of available texts. Notably, this sampling is meticulously tailored to the content of the questions themselves, emphasizing a targeted selection process that aligns the model's training data with the specific challenges posed by the competition.

\section{Related Works}

Visual Question Answering (VQA) can be divided into three main categories. The first deals with realistic or synthetic images with the task of associating visual parts in the scene with texts or recognizing objects. The second category deals with analyzing scientific graphs with the aim of deriving properties such as trends. The last area, the one of interest in this work, focuses on the processing of scanned documents to retrieve relevant information, like sections linked to queries.

Regarding the latter category, it can focus on two different document types: single-page or multi-page documents. In a single-page context, the best models available to date vary from Document Understanding techniques such as~\cite{docformer2, donut, ernie, layoutlmv2, layoutlmv3}, Visual Language Understanding techniques~\cite{pix2struct} or Large Models~\cite{pali}.
On the other hand, for models dealing with multi-page processing, the current state of the art is very limited compared to the other case study as it is a recent research topic.
In fact, many proposed solutions are based on word concatenations before applying single-page strategies thus not scalable in a multi-page scenario,
while others propose hierarchical and graph-based approaches to tackle the multi-page domain~\cite{hi,pdfvqa}.


\section{Methodology}

\subsection{Task and Dataset}

This challenge encapsulates a Visual Question Answering (VQA) task, a multifaceted endeavor that demands the fusion of natural language comprehension with visual document analysis. The objective is to develop a model that, given a natural language question, can accurately pinpoint the Region(s) of Interest (RoIs) which provides the correct answer to the question.
These RoIs encompass titles, textual content, images, and tables, constituting the vital components for addressing the posed questions.

The provided dataset, denominated as PDF-VQA\cite{pdfvqa}, comprises a rich collection of 1147 multi-page documents meticulously sourced from PubMed, properly divided into 800, 115, and 232 for train, validation, and test sets respectively. Crucially, the dataset is pre-processed to provide the RoIs, the textual component of each ROI, and some information regarding the relationship between document components, such as spatial and parent-child relationships. Accompanying the document corpus is a set of 5653 questions divided into 3951, 581 and 1121 corresponding to the train, validation and test sets, driving the model to engage with a wide spectrum of inquiries. Questions are divided into two macro-categories: parent-relationship (PR) understanding (3151 of train, 466 of validation, 889 of test) and child-relationship (CR) understanding (800 of train, 115 of validation, 232 of test). 

Given the predominance of PR questions, we further investigated their content. Upon thorough analysis, we observe that a significant majority of PR questions are directly related to figures and tables. Specifically, within the train split 2396 out of 3951 questions, in the validation split 359 out of 581, and in the test split 675 out of 1121 were centered around these critical document elements. This emphasis further underscores the importance of our tailored approach, which leverages this insight to hone the model's focus on the most frequently queried components.

\subsection{Method}

Our proposed methodology hinges on a two-step process: fine-tuning a BERT model and leveraging the resulting embeddings for decoding answers to the posed questions.

\subsubsection{Fine-tuning with Keyword-Enriched Texts}

To impart domain-specific knowledge to our model, we meticulously select text snippets from the training documents that contain explicit references to figures or tables, or their abbreviated forms. In detail, our selections reduce the training samples from 75791 
to 10543.
This curated subset not only ensures that our model focuses on the pivotal elements that are most frequently queried but also significantly reduces the training time.
 
A BERT model is then fine-tuned using the Masked Language Modeling (MLM) task, a technique that has proven highly effective in a myriad of NLP applications. Adhering to the original BERT paper's settings\cite{bert}, we allocate 15\% of the total tokens within each sentence for transformation. 
When a token is selected we apply masking 80\% of the time, we replace it with a random token from the vocabulary 10\% of the time and we leave it unchanged in the remaining 10\% of the time. 
The loss is finally computed taking into account all selected tokens.

\subsubsection{Embedding Extraction and Decoding}
We use our fine-tuned model to extract textual embeddings from the entire training corpus. We feed the embeddings into a transformer-based decoder, coupled with the posed question, to complete the task. This multi-step process capitalizes on the enriched contextual understanding afforded by our BERT model, enabling our system to discern and provide accurate answers across the diverse array of questions posed.

\section{Experimental Results}

This section furnishes comprehensive insights into the methodological particulars that underpinned our experimental design. It also encapsulates the substantive outcomes and observations garnered through our analyses. 

\subsection{Settings}
Regarding BERT's Masked Language Modeling training, CrossEntropy loss combined with AdamW as an optimizer with a learning rate set to $1\cdot10^{-4}$ and weight decay set to $0.01$ were used. Finally, the model was finetuned for $50$ epochs. 
Concerning the setting of the final model, we took over the structure provided in the baseline for the VQA model. 
In detail, we use BCE loss, Adam optimizer with a learning rate set to $5\cdot10^{-5}$. We also set the number of decoder layers to $5$, and we train the model for $30$ epochs.
For both models, the best checkpoints based on validation loss
were taken and used, while, regarding the hardware resources, all experiments were conducted on a system equipped with a Nvidia RTX A6000.

\subsection{Results}
We evaluate system performance in terms of \textit{exact matching accuracy} (EMA), i.e., the number of correct question-answer matching over all the questions. 
Our proposed approach demonstrated significant improvement, achieving a score of $0.35491$ on the public leaderboard and a score of $0.38484$ on the private one, concluding with a second place overall in the competition. This marked enhancement surpasses our baseline result of $0.32142$, which was obtained using textual embeddings extracted with the BERT model without any model specialization. 

Interestingly, our text-centric approach outperforms its multimodal counterpart. When compared to a method involving the concatenation of visual features extracted with ResNet-152 to the textual embeddings before feeding them to the decoder, our proposed approach stands out as the more effective strategy.

We explored alternative in-domain fine-tuning, such as contrastive fine-tuning of the CLIP model~\cite{clip} using figures and tables along with their captions and contrastive finetuning of BERT using sentences with relevant keywords as positive samples and sentences without relevant keywords as negative ones. 
However, it was the MLM task with the meticulously reduced subset that ultimately yielded the most promising results. 

\section{Conclusions and Future works}
The Visual Question Answering task within the multi-page domain is a complex challenge. Our proposed methodology, rooted in fine-tuning a BERT model on a data sampling, has yielded substantial accuracy improvements. The proposed strategy outperforms the multimodal variant and other in-domain fine-tuning methods. 

However, some considerations can be made about our solution. 
Our proposal is mainly based on the Parent-Relationship (PR) understanding sub-task, with a special focus on keyword matching between questions and texts. This strategy can be enhanced both by extending the set of relevant keywords to other PRs, such as in-text citations, and by proposing a fine-tuning subset of data tailored to the Child-Relationship (CR) understanding sub-task. 

In prospective endeavors, our intention is to persist in the refinement of our proposal, with the overarching objective of enhancing its performance. This pursuit will encompass the exploration of different methodologies for amalgamating visual features, as well as the strategic utilization of alternative representations predicated upon the intrinsic relationships existing between constituent elements of each document.


\bibliographystyle{ACM-Reference-Format}
\bibliography{sample-sigconf_FINAL}

\end{document}